\documentclass[journal,twocolumn]{IEEEtran}

\usepackage{amsfonts}
\usepackage{times}
\usepackage{graphicx}
\usepackage{latexsym}
\usepackage{dsfont}
\usepackage{amssymb}
\usepackage{amsmath}
\usepackage{cite}
\usepackage{verbatim}

\newcommand{\figref}[1]{{Fig.}~\ref{#1}}

\def\bb0{{\mathbb{0}}}

\def\bb{{\mathbf{b}}}

\def\b0{{\mathbf{0}}}

\def\sf0{{\mathsf{0}}}

\usepackage{epstopdf}
\usepackage{enumerate}
\usepackage{algorithmicx}
\usepackage{algorithm}
\usepackage{amsmath}
\usepackage[noend]{algpseudocode}
\usepackage{float}
\usepackage{hyperref}
\usepackage{color}
\usepackage{makeidx}
\usepackage{subcaption}
\usepackage{bbm}
\usepackage{graphicx}
\usepackage{lipsum}
\usepackage{tablefootnote}
\usepackage{soul}
\usepackage{multirow}
\usepackage{multicol}
\usepackage{balance}

\newcommand{\sref}[1]{{Section}~\ref{#1}}

\begin{document}
	
	\title{Computer Vision Aided URLL Communications: \\ Proactive Service Identification  and  Coexistence }
	\author{Muhammad Alrabeiah, Umut Demirhan, Andrew Hredzak, and Ahmed Alkhateeb \\ Arizona State University, email: $\{$malrabei, udemirhan, ahredzak, alkhateeb$\}$@asu.edu}
	\maketitle
	
\begin{abstract}
The support of coexisting ultra-reliable  and low-latency (URLL) and enhanced Mobile BroadBand (eMBB) services is a key challenge for the current and future wireless communication networks. Those two types of services introduce strict, and in some time conflicting, resource allocation requirements that may result in a power-struggle between reliability, latency, and resource utilization in wireless networks. The difficulty in addressing that challenge could be traced back to the predominant reactive approach in allocating the wireless resources. This allocation operation is carried out based on received service requests and global network statistics, which may not incorporate a sense of \textit{proaction}. Therefore, this paper proposes a novel framework termed \textit{service identification} to develop novel proactive resource allocation algorithms. The developed framework is based on visual data (captured for example by RGB cameras) and deep learning (e.g., deep neural networks). The ultimate objective of this framework  is to equip future wireless networks with the ability to analyze user behavior, anticipate incoming services, and perform proactive resource allocation. To demonstrate the potential of the proposed framework, a wireless network scenario with two coexisting URLL and eMBB services is considered, and two deep learning algorithms are designed to utilize RGB video frames and predict incoming service type and its request time. An evaluation dataset based on the considered scenario is developed and used to evaluate the performance of the two algorithms. The results confirm the anticipated value of proaction to wireless networks; the proposed models enable efficient network performance ensuring  more than $85\%$ utilization of the network resources at $\sim 98\%$ reliability.  This highlights a promising direction for the future vision-aided wireless communication networks.
\end{abstract}
	

 \section{Introduction}\label{sec:intro}
A defining feature for future wireless communication networks is the ability to support a wide spectrum of heterogenous coexisting applications, from Machine-Type Communications (MTC) and Mission-Critical Communications (MCC) to the rate-hungry enhanced-Mobile BroadBand (eMBB) \cite{URLLC_Bennis2019,Azari2019}. A spectrum like that results in wireless networks that are riddled with many challenges \cite{Saad2020}; on the one hand, they need to satisfy the growing data rate demands of eMBB applications, and on the other, they must satisfy the stringent constraints of Ultra Reliability and Low Latency Communications (URLLC) that is demanded by MCC and MTC applications. Dealing with challenges like those requires a paradigm shift in designing wireless networks, moving away from reactive networks towards proactive ones \cite{URLLC_Bennis2019}.

The key word in the description of that paradigm shift is ``proactive;'' it points to a clear need for a network that can anticipate incoming services and incorporate that in its resource allocation. To have such capability, a wireless network needs to have a sense of awareness; it should be able to observe its surrounding, make sense of what it has observed, and utilize its knowledge of the surrounding to enable proactive resource allocation. This awareness could be accomplished with machine learning and more specifically with its most advanced learning paradigm, deep learning \cite{DLBook}. This is confirmed by recent studies like \cite{Saad2019,Azari2019}. They make the case for machine learning and emphasize its role in future wireless networks.

The ability of a machine learning algorithm to learn a task is, in some sense, upper bounded by the amount of information the observed variables--and desired responses in case of supervised learning--convey about the task itself\footnote{This is a direct consequence of the data processing inequality \cite{InfoTheory} where the machine learning algorithm could be viewed as a multi-layer processing operation.}. As such, choosing the right type of data modality to enable awareness is very critical. For a wireless network, visual data, like images and videos, represent a great and untapped resource of information. Visual data is rich with clues about the environment where the wireless network operates, and with the right machine learning algorithm, relevant information could be learned to facilitate awareness and proaction.

In this work, computer vision, which is the field where machine learning and visual data merge, will be used to enable proactive resource allocation in wireless networks. More specifically, a framework termed \textit{service identification} is proposed as a new layer of intelligent processing that should aid resource allocation. It is centered around the idea of identifying \textit{incoming} services using multimodality data that includes visual data like RGB images, video frames,...etc. This framework is envisioned as an enhancement of existing resource allocation strategies, especially those focused on proaction. The following subsection will summarize the relevant work in the literature before the main contributions of this paper are detailed in the next one. 

\subsection{Prior Work}
A survey on the challenging requirements of URLLC, recent advances and potential applications can be found in \cite{sutton2019enabling}. Specifically, the optimization of the resources for URLLC require different statistical approaches than those considered for standard rate-centered communications; reliability and latency need to be jointly considered in addition to standard data rate maximization \cite{URLLC_Bennis2019}. For enabling URLLC, different statistical tools, protocols and access methods are investigated in a significant amount of work (e.g., \cite{popovski2019wireless}) with growing recent interest in the coexistence of eMBB and URLLC. Good examples could be found in \cite{alsenwi2019embb, anand2020joint, bairagi2020coexistence}. They focus on proposing puncturing/scheduling solutions for mixed eMBB and URLLC traffic. A common drawback across all of them is their reactive nature; the scheduling only starts when a request is made.

Another line of work that has recently been emerging relies on utilizing machine learning to address the coexistence and scheduling of URLL and eMBB traffic, especially for cases with unscheduled URLLC traffic.\cite{azari2019risk, elsayed2019ai, li2020deep} propose machine learning aided approaches that proactively allocate resources and facilitate URLL communications. More specifically,  \cite{azari2019risk} proposes a risk-aware online learning approach for the allocation of resources, and \cite{elsayed2019ai} proposes Q-learning for the allocation of power and time-frequency resources. Similarly, the authors in \cite{li2020deep} utilize a model-free deep reinforcement learning algorithm, called deep deterministic policy gradients, for the allocation of resource blocks allowing the URLL traffic. All that work is intriguing as it touches upon the concept of proaction in wireless networks; however, what it is collectively lacking is the utilization of rich sources of information, especially visual data.

\subsection{Contribution}
This paper introduces the service identification framework for future wireless networks. As mentioned earlier, in its core is deep learning and computer vision, and its ultimate objective is to provide the wireless network with the sense of awareness it needs to achieve proaction. The following three points provide a break-down of the main contributions:
\begin{itemize}
  \item Service identification is proposed as a two-component framework, in which not only service type but also its request time (i.e., when the service is starting) are predicted ahead of time. Such prediction is based on, but not limited to, an observed video sequence from the environment where the wireless network operates.
  \item For the sake of illustrating the potential of this framework, two deep learning algorithms based on Deep Neural Networks (DNNs) are proposed to predict service type and request time. The two algorithms are evaluated in a spectrum-sharing scenario where the wireless network aims at meeting the requirements of two different services, eMBB and URLL.
  \item A synthetic indoor scenario is develop using the ViWi data-generation framework \cite{ViWi}. This scenario is used to generate the development and evaluation dataset for the proposed two DNNs. 
\end{itemize}

The proposed framework and deep learning algorithms are detailed in the following six sections. In particular, Section \ref{sec:key_idea} will start off with a high-level description of the proposed framework, emphasizing its key idea and how it could be utilized in future wireless networks. Following that, Section \ref{sec:sys_prob} provides a formal description of the system model and the service identification problem definition adopted in this work. In Section \ref{sec:prop_sol}, two DNN architectures are proposed as candidate solutions to the service identification problem. In order to study their performance, Section \ref{sec:exp_setup} presents a description of the experimental setup, which includes a description of the development and evaluation dataset, the DNNs training, and the performance-evaluation metrics. The actual evaluation of their performance will proceed in Section \ref{sec:perf_eval}, and, in the end, Section \ref{sec:concl} will wrap up this paper with a summary of its main findings.  

\begin{figure*}[ht]
\centering
	\includegraphics[width=0.8\linewidth]{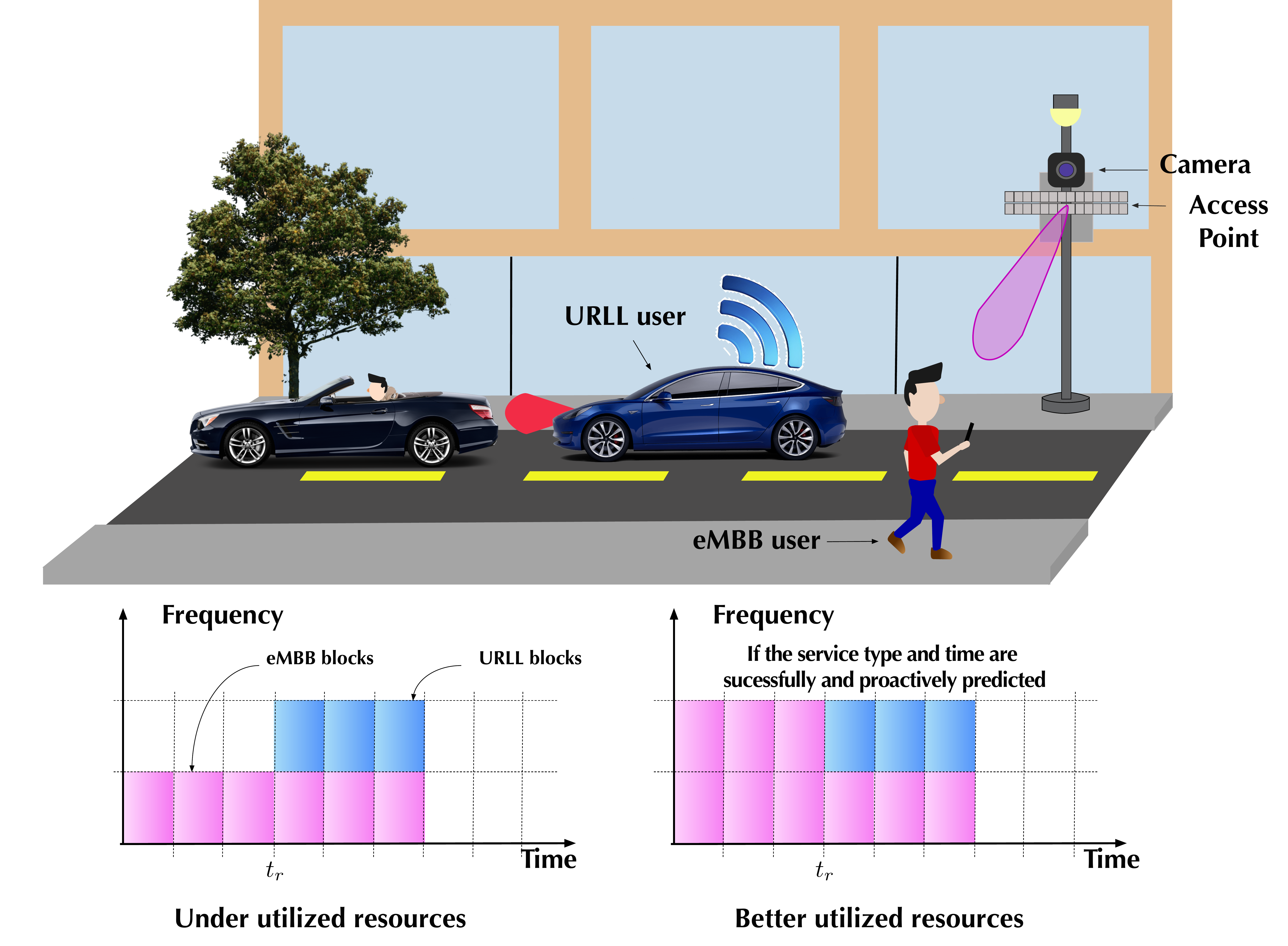}
	\caption{An example illustrating the key idea behind the service identification framework. It shows an access point with a camera carting to two types of services, URLL and eMBB. The bottom part of the figure also depicts a contrast between two different resource allocation schemes the access point may adopt}
	\label{fig:key_idea}
\end{figure*}

\section{Vision-Aided Service Identification: Key Idea}
\label{sec:key_idea}

The need to support heterogeneous coexisting services in future wireless networks results in an overwhelming strain in terms of reliability, spectral efficiency, and latency \cite{Saad2020}. Allocating resources to meet the wide range of demands that come with those services becomes an elusive task when addressed using modern-day approaches to resource management. This is due to the fact that the wireless network operation is majorly reactive \cite{URLLC_Bennis2019} and it lacks agility as the resources are allocated upon request. 

Proactive resource allocation is one promising way to alleviate the strain arising from heterogeneous coexisting services; it enables non-orthogonal resource allocation and provides the network the agility it needs to meet the different requirements. However, an intuitive question one could pose at this point is: \textit{How could a wireless network achieve proaction?} The answer to this question could be rooted in the ability of the wireless network to be aware of the surrounding environment (including the user behavior). \textbf{In particular, in order to develop a sense of environment awareness, a network is envisioned to have the ability to \textit{monitor} its surrounding, \textit{detect} elements of interest, and, finally, \textit{reason} the implications of the behavior of those elements in order to make decisions.} This three-stage process (i.e., monitor, detect, and reason) hints at the need for an intelligent wireless network, in the core of which are two important ingredients: (i) a rich source of information about the environment, and (ii) an intelligent algorithm that makes use of the information source and proactively manages the resources.

Those two ingredients could be found in visual data (e.g., RGB images, depth maps, and video sequences) and deep learning. The former is a very rich source of information while the latter has been driving the state-of-the-art in machine learning and computer vision \cite{DLBook,DL:meth_App}. With visual data, a wireless network has what it needs to monitor the environment and how it changes, which is the first stage in the way to proaction. This data is analyzed using a deep learning algorithm to identify relevant objects and understand their behavior. The algorithm learns to anticipate incoming wireless services based on its understanding of the objects in the environment and their behaviors, which basically represents the last two stages to proaction, namely detection and reasoning.

\subsection{Key Idea in Context}\label{sec:key_idea_cont}
The above intuition about service identification could be put into the context of resource allocation through an example. Consider a communication environment like that in \figref{fig:key_idea}. It depicts a dynamic outdoor wireless environment where a basestation ( or access point) serves two different types of applications, URLL and eMBB. In classic operation settings (no vision or machine learning exist), the network may use an orthogonal resource slicing such as that presented in the bottom left time-frequency plot of \figref{fig:key_idea}. This slicing result in high reliability because each service is guaranteed an allotted resource block over some time. However, as expected, it severely under-utilizes the resources. 

In this paper, we propose to leverage computer vision to develop  a more agile and efficient recourse allocation approach that guarantees both high utilization and high reliability. \textbf{By equipping the base station in \figref{fig:key_idea} with a camera and a powerful machine learning algorithm, the network is enabled to \textit{monitor, detect, and reason} from the context of its surrounding environment}. More to the point, it can identify the existence of two possible objects of interest, a pedestrian on their phone and an autonomous vehicle approaching another vehicle. Based on the behavior of the two objects, the former could be anticipated to be requesting an eMBB service while the latter is more likely to request a URLL service very soon. Such understanding can trigger a smart vision-aided resource allocation scheme; based on the speed and distance between the two vehicles, the wireless network may anticipate the time at which a URLL service is going to be requested. Therefore, it could allow the eMBB service to utilize all available resource blocks until that anticipated time, at which it scales down the eMBB utilization to accommodate the incoming URLL. In an ideal setting, vision-aided resource allocation could result in an allocation scheme similar to that depicted in the bottom right time-frequency plot of \figref{fig:key_idea}.

\subsection{Definition of Service Identification}\label{sec:key_idea_def}
Form the intuition developed at the beginning of this section as well as the example scenario in \figref{fig:key_idea}, the framework of service identification could be defined as:
\begin{quote}
	a vision-aided multi-modal learning framework in which one or multiple machine learning algorithms are designed to induce proaction in wireless communication networks. This proaction manifests in two fundamental tasks: (i) predicting the types of incoming services, and (ii) predicting the request times of those services, i.e., the times at which those incoming services will be requested.
\end{quote}
The above definition has four landscape-defining phrases. The first two are ``vision-aided multi-modal'' and ``machine learning algorithms;'' these two represent the main ingredients for an intelligent wireless network, which are the rich source of information and the intelligent algorithm. They, as discussed earlier in this section, hold an answer to how proaction could be achieved in wireless networks. The other two phrases, namely ``predicting the types'' and ``predicting the request times,'' emphasize the two main tasks lying in the core of the framework. They represent the major two tasks the intelligent algorithm needs to perform based on what it learns from the data the information source provides. The rest of this paper is devoted to showing how the above definition could be translated into a solution that improves reliability and spectral utilization in a typical wireless network.

\section{System Model and Problem Definition}\label{sec:sys_prob}
In the following two subsections, we present the adopted system model and problem definition. 
\subsection{System Model}\label{sec:sys_mod}
\begin{figure}[t]
	\centering 
	\includegraphics[width=1\linewidth]{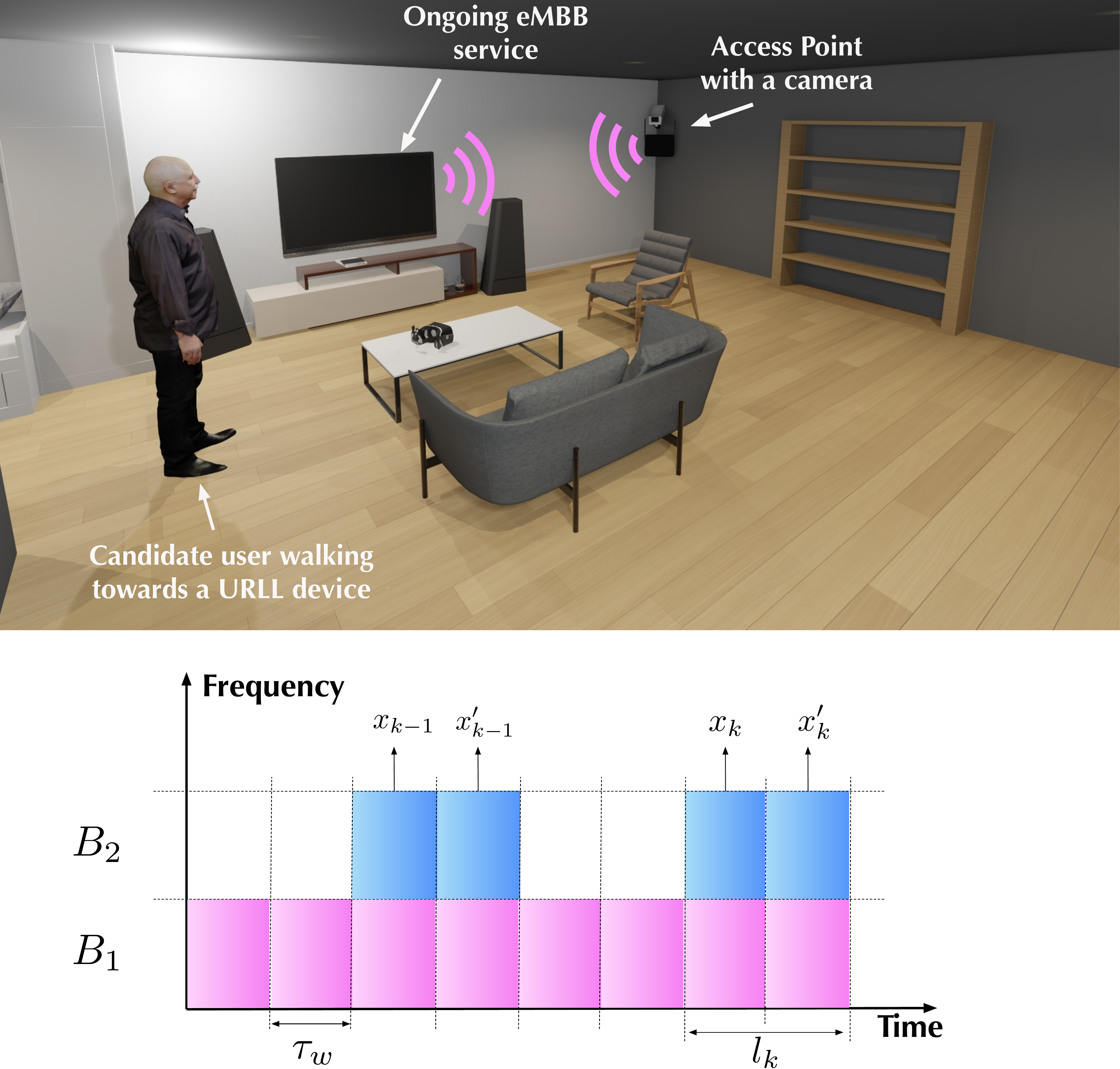}
	\vspace{5pt} 
	\caption{An illustration of the adopted communication scenario and the resource blocks available for the wireless network.}
	\label{fig:sys_model}
\end{figure}

We consider the scenario depicted in \figref{fig:sys_model} where a single access point (AP) equipped with an RGB camera serves coexisting  eMBB and URLLC services. The available frequency band is assumed to be divided into two sub-bands $B_1$ and $B_2$. The eMBB service can use both $B_1$ and $B_2$. It will first, though, utilize $B_1$ and it will start utilizing $B_2$ only if $B_1$ is already utilized. The URLLC can only use $B_2$. Further, to satisfy the low-latency requirements, the URLLC packets are transmitted directly over $B_2$, without any initial-access or hand-shaking protocol. This means that when the eMBB service is provided through both $B_1$ and $B_2$, an incoming URLLC transmission observes interference and this consequently results in the loss of the URLL packet. 

In addition to the available frequency sub-bands, our system model assumes that the time in split into time slots; each of duration $\tau_w$. In other words, each resource block has a frequency bandwidth $B_1$ or $B_2$ and a time duration $\tau_w$. Without loss of generality, we assume that the  eMBB service is always operating and is always utilizing $B_1$.  For the URLLC service, we assume that they can start randomly at any time slot and the transmission continues until it transmits all of its data. We will elaborate more on how the URLLC service starts in Sections \ref{sec:prop_sol} and \ref{sec:exp_setup}.

\begin{figure}
	\centering
	\includegraphics[width=\linewidth]{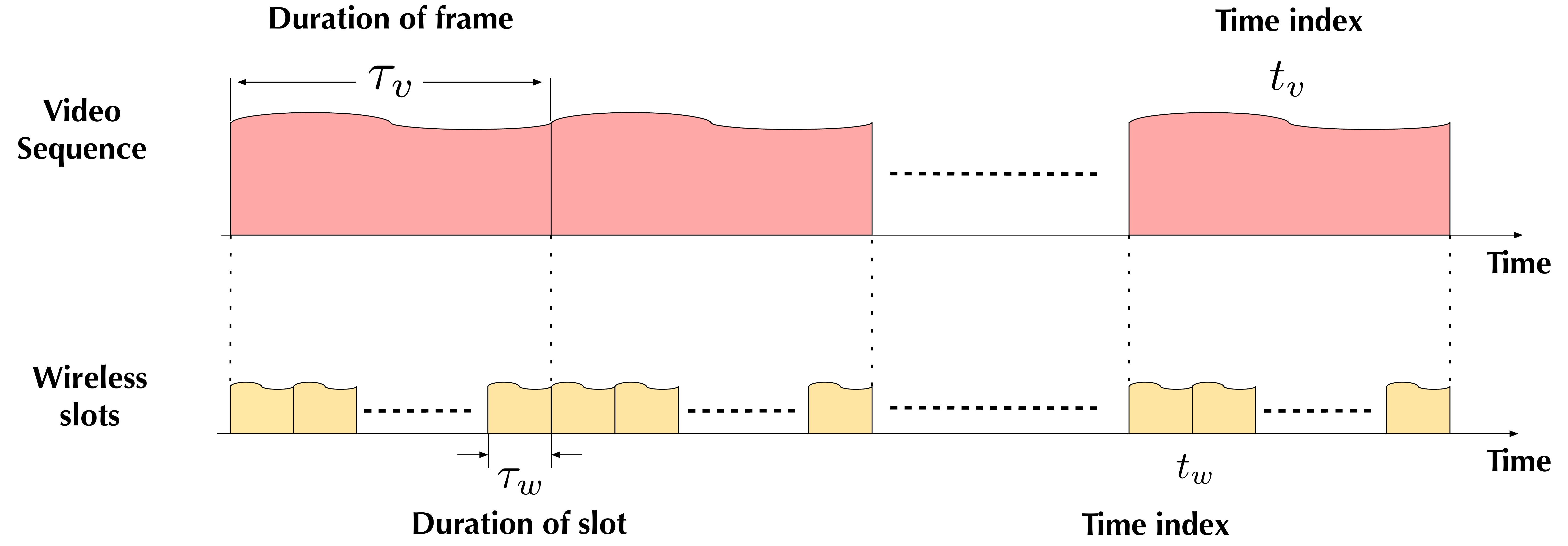}
	\caption{The figure illustrates the relation between duration and time index of the video frames and the wireless resource slots.}
	\label{fig:wi_vi_time}
\end{figure}

The AP also relies on the RGB camera to provide a continuous stream of video frames. This video depicts the various objects the make up the wireless environment. The camera is running at a frame rate of $1/\tau_v$ frames/second, where $\tau_v$ defines the duration of a single video frame. Given the fact that commodity video cameras run at frame rates between 30 to 120 frames/second, the duration of a video frame is consistently larger than a wireless slot (i.e., $\tau_w \leq \tau_v$), which is usually in the order of a fraction of a millisecond \cite{5g_overview}. To characterize the relation between the two, a vision-aided wireless network defines the Frame-to-Slot Ratio as $\text{FSR}=\tau_v/\tau_w$. This ratio is critical to any joint processing of wireless and visual information, which is the bedrock of the service identification framework.
 For the sake of simplicity in this paper, the video frames and wireless slots are assumed to be synchronized such that the beginning of a video frame corresponds to the beginning of a wireless slot, see \figref{fig:wi_vi_time}.

\subsection{Problem Formulation}\label{sec:prob}

With the system model above in mind, we now define the problem from a wireless network perspective and, then, translate it into machine learning terms. Let us denote the slots where the $k$-th URLL packet starts and ends by $x_k$ and $x^{\prime}_k$, respectively. We denote the length of the packet as $l_k = x^{\prime}_k-x_k+1$. The variable $s_k$ is defined as the indicator of the success of the $k$-th URLL packet. To clarify, $s_k=0$ if the packet collides with the eMBB transmissions, and $s_k=1$ if there is no collision. We define the \textit{reliability} $R$ as the percentage of the successful URLL transmissions. After the transmission of $K$ URLL packets, it can be written as
\begin{equation}\label{eq:rel}
    R\, \% = \frac{\sum_{k=1}^K s_k}{K} \times 100.
\end{equation}

Next, we define a variable $z_q$ that indicates the utilization of $q$-th slot. $z_q=1$ if there is a transmission in the $q$-th slot of the URLL bandwidth, and $z_q=0$ when there is no transmission in the bandwidth. We note that the utilization may be due to the URLL packets as well eMBB transmissions. Then, we define \textit{utilization} $Z$ as the number of slots that are being used by either of the services. After the completion of the wireless slot $t_w$, we can write 
\begin{equation}\label{eq:eff}
    Z\% = \frac{\sum_{q=1}^{t_w} z_q}{t_w} \times 100.
\end{equation}

\begin{figure*}[t]
	\centering
	\begin{subfigure}{.8\textwidth}
		\centering
		\includegraphics[width=\linewidth]{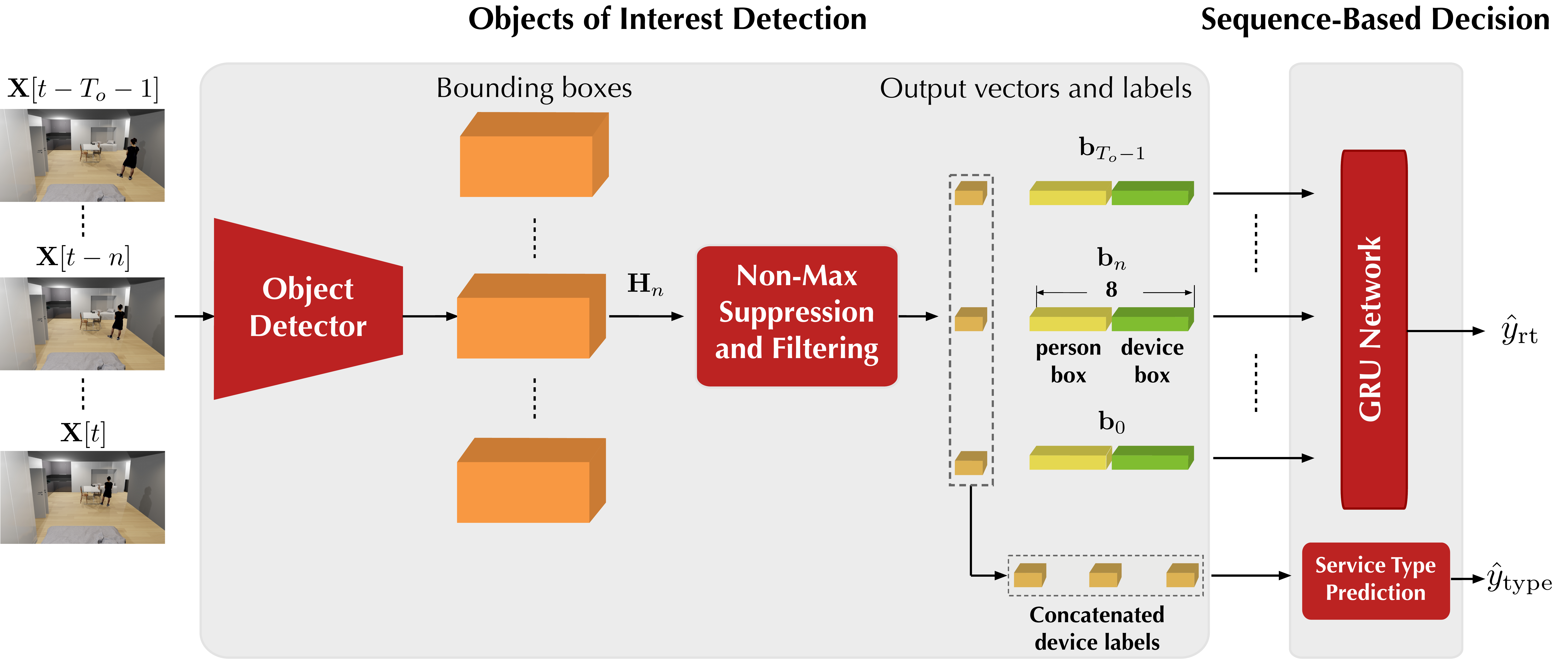}
		\caption{ The Two Stage Architecture}
		\label{fig:2_stage}
	\end{subfigure}
	\quad
	\begin{subfigure}{.8\textwidth}
		\centering
		\includegraphics[width=1\linewidth]{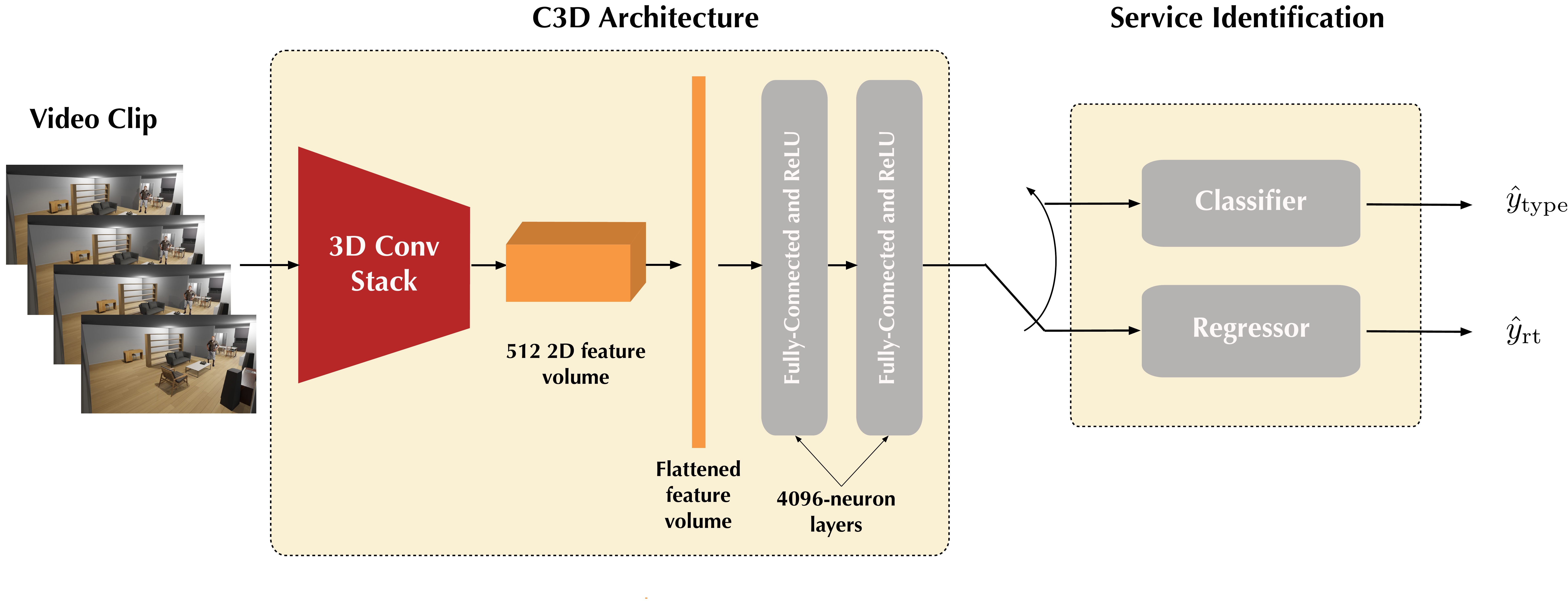}
		\caption{ The End-to-End Architecture}
		\label{fig:end_end}
	\end{subfigure}
	\caption{This figure presents the two schematics of the proposed DNN architectures. (a) shows the two-stage architectures while (b) shows the end-to-end architecture. The two architectures are described in detail in \sref{sec:prop_sol}.}
\end{figure*}
 
Due to the coexistence of the services, the defined metrics, reliability and utilization, are closely related. For instance, perfect (i.e., 100\%) reliability can be achieved when the URLL bandwidth is always kept empty by the eMBB service. However, such an approach would result in degradation in the data rate of the eMBB and the utilization rate. Therefore, to facilitate the coexistence of the services in an efficient manner and achieve high utilization rates, we seek a solution based on proactive service identification that could provide high data rates for the eMBB service while satisfying a desired reliability for the URLL service.

Given the definition of the service identification framework in Section \ref{sec:key_idea_def}, the problem definition above could be translated into machine learning terms as follows. Assume an observation time window of length $T_o$ instances, and let $\mathbf X[t_v-n] \in\mathbb R^{W\times H\times C}$ be an RGB video frame captured at the $(t_v-n)$-th time instance of that window, where $t_v\in\mathbb Z$ is a video time index,  $n\in\{0,\dots,T_o-1\}$, and $W$, $H$, and $C$ are, respectively, the width, height, and number of color channels of the image. In addition, define $f^{(1)}_{\Theta_1}(\mathcal S_{o})$ and $f^{(2)}_{\Theta_2}(\mathcal S_{o})$ to be two prediction functions parameterized by the two sets of trainable parameters $\Theta_1$ and $\Theta_2$. Both take on a sequence of observed images $\mathcal S_{o}=\{X[t_v-n]\}_{n={T_o-1}}^{0}$ and produce predictions on, respectively, the type of service $\hat y_{\text{type}}\in\{0,1\}$ and its request time $\hat y_{\text{rt}}\in\mathbb R$, where 0 refers to an eMBB service and 1 refers to an URLL service. \textbf{Then, the objective of the proactive service identification in this paper is to learn the two functions $f^{(1)}_{\Theta_1}(\mathcal S_{o})$ and $f^{(2)}_{\Theta_2}(\mathcal S_{0})$ such that they maximize the probability of correct detection for both service type $\mathbb P(\hat y_{\text{type}} = y_{\text{type}} | \mathcal S_o)$ and request time $\mathbb P(\hat y_{\text{rt}} = y_{\text{rt}} | \mathcal S_o)$}. It is important to note here that if the service type and request time can be perfectly predicted, this means achieving optimal (100\%) reliability and utilization, which is the overarching goal of the proposed vision-aided  proactive service identification approach.

\section{Proposed Deep Leaning Solution}\label{sec:prop_sol}
Prediction of service type and request time (i.e., service identification) could be viewed as action recognition tasks in the context of wireless networks, for which advances in DNNs \cite{DLBook} and video analytics \cite{3DConv, r-c3d, QuoVadis} can be utilized. In this paper, two DNNs architectures for service identification are proposed based on two different design perspectives.
One tackles the problem from the angle of ``two-stage'' learning while the other does that from the angle of ``end-to-end'' learning. The details of the two are given in the following two subsections.

\subsection{Two-Stage Neural Network Architecture}\label{sec:2_stage}
This architecture, as it name suggests, divides the learning process into two stages, namely objects of interest detection and sequence-based decision, see \figref{fig:2_stage}. The first stage employs an object detector network that is tasked with identifying objects of interest in an observed video sequence $\mathcal S_{o}$ capturing the wireless environment. Hence, it is trained to detect different classes of objects in each frame of $\mathcal S_{o}$, e.g., detect people, cars, phones, computers, and so on. For each RGB frame, the outputs of the detector are organized into a matrix $\mathbf H_{n} = [\mathbf h_{1},\dots,\mathbf h_{M_n}]^T\in\mathbb R^{M_{n}\times6}$ where $M_{n}$ is the number of objects detected in $n$-th frame and 6 defines the dimensions of the object vector $\mathbf h$. This vector contains the 4 pixel coordinates marking the top-left and bottom-right corners of the bounding box surrounding the detected object; the maximum class probability of the object; and, finally, the confidence of the detector in the bounding box having an object of interest. 

Each matrix $\mathbf H_n$ is fed to the non-maximum suppression and filtering component in the first stage. This component, first, eliminates any redundant object vectors using class probability and detector confidence. Then, it filters the surviving object vectors using object confidence to pick only two vectors. One represents a candidate user $\mathbf h_{\text{user}}$ and the other represents a candidate device $\mathbf h_{\text{device}}$. Finally, this component transforms the two object vectors into an output vector $\mathbf b_n\in\mathbb R^{8\times1}$ and a label, see \figref{fig:2_stage}. The vector $\mathbf b_n$ is a concatenation of the bounding boxes coordinates in $\mathbf h_{\text{user}}$ and $\mathbf h_{\text{device}}$ whereas the label is the type of device that has been detected in the current frame. This output vector and label are passed to the second stage in this architecture. 

The second stage takes in $T_o$ output vectors $\mathbf b_n,\ n\in\{1,\dots,T_o\}$, and $T_o$ labels to make the predictions on the type of service and its request time. The output vectors are fed to a recurrent neural network composed of two layers of Gated Recurrent Units (GRUs). This network essentially learns how the distance between the device and the user is changing over frames, and it uses that to predicted at which time in the future the service will be requested. On the other hand, to determine the type of that service, the $T_o$ labels are concatenated in one vector and a majority voting rule is used; the device type that repeats more in the object detector predictions determines the type of service as it reflects a sense of temporal sense of confidence in the detector predictions--something that cannot be done using single image detection.

\subsection{End-to-End Neural Network Architecture}\label{sec:end_to_end}
The second proposed architecture takes a holistic approach to learning the two tasks of service identification. The basic idea behind its design is founded in the utilization of spatiotemporal information present in a video sequence. This is accomplished using 3D convolution and 3D max pooling layers, which have proved effective in learning action recognition tasks \cite{3DConv,QuoVadis,r-c3d}. For the problem in hand, the popular C3D network \cite{3DConv}  is used as the base network of the proposed architecture, see \figref{fig:end_end}. Since it is originally designed for 16-frame long input videos, the network is tweaked across the the time dimension to fit the service identification problem, in which the video has $T_o$ frames. In particular, the number of 3D convolution and 3D max pooling layers is kept the same as that in the original network (5 layers each), yet the sizes of their kernels are modified such that the important information in the input $T_o$ frames is squeezed into 512 2D feature maps. These maps are then flattened into high-dimensional vector and fed to two 4096-neuron stacks of fully-connected and ReLU \cite{DLBook} layers, which are expected to learn task specific features. More details on the architecture and its implementation could be found at \cite{myGithub}. 

To make predictions on the service type and request time, this architecture has two different and independent prediction layers. Service type is predicted using a classifier while request time is predicted using a regressor. The classifier is built with a two-neuron fully-connected layer followed by a softmax activation \cite{DLBook}, and the regressor is build with a single-neuron fully-connected layer, see \figref{fig:end_end}. What is important to note here is that the architecture is trained independently for each task. This is done by switching the last layer and training the whole network end-to-end. This results in two trained models that only differ in the last layer.

\section{Experimental Setup}\label{sec:exp_setup}
Since the proposed solutions are machine learning based, a development dataset and a set of evaluation metrics need to be developed to assess their effectiveness in performing service identification and proactive resource allocation. The following few subsections will introduce those elements and discuss how each proposed DNN is trained.

\begin{figure}
	\centering
	\begin{subfigure}{0.5\textwidth}
	    \centering
		\includegraphics[width=1\columnwidth]{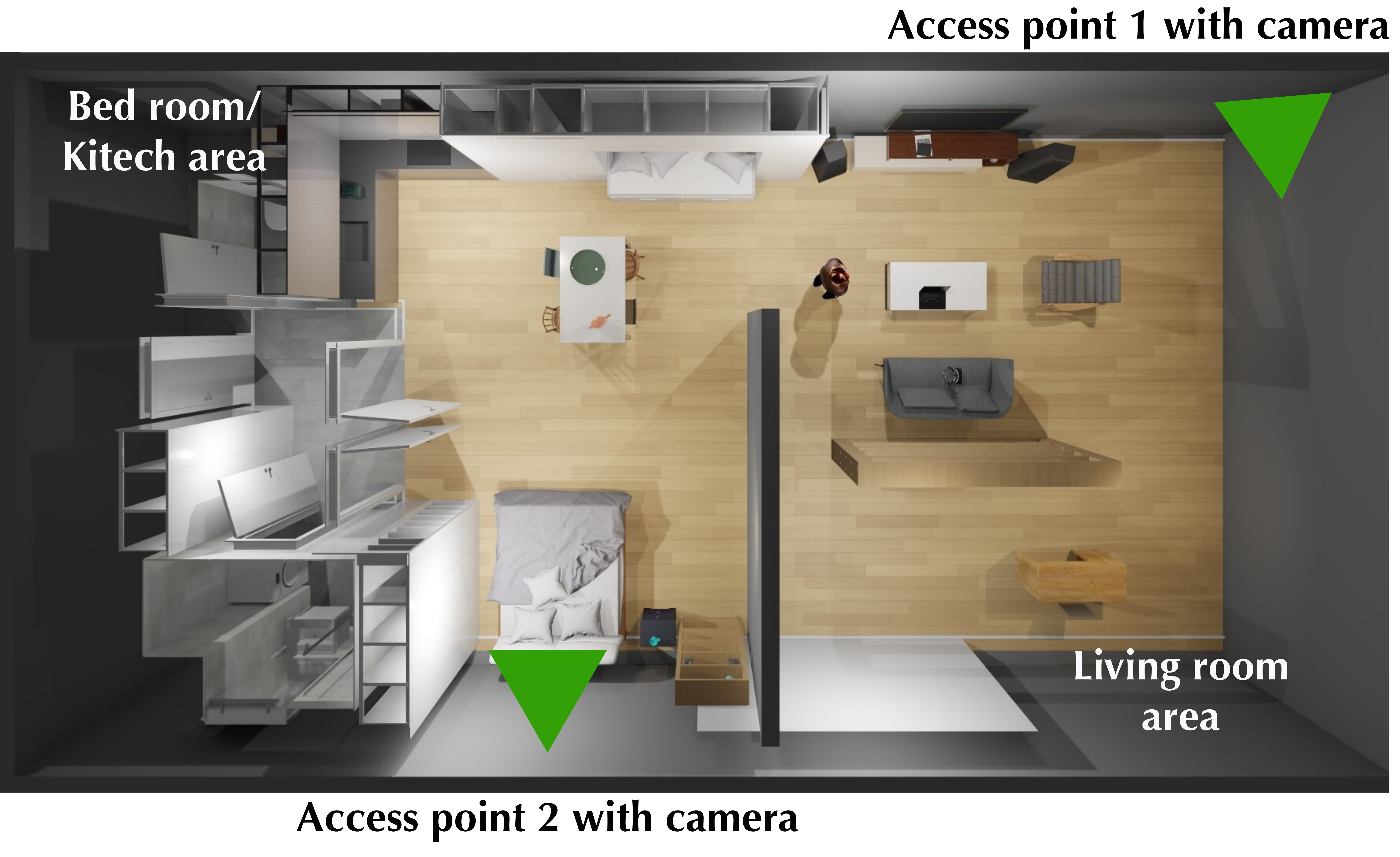}
		\caption{Top-view}
		\label{fig:scen}
	\end{subfigure}
	\par\bigskip
	\begin{subfigure}{0.5\textwidth}
	    \centering
		\includegraphics[width=1\columnwidth]{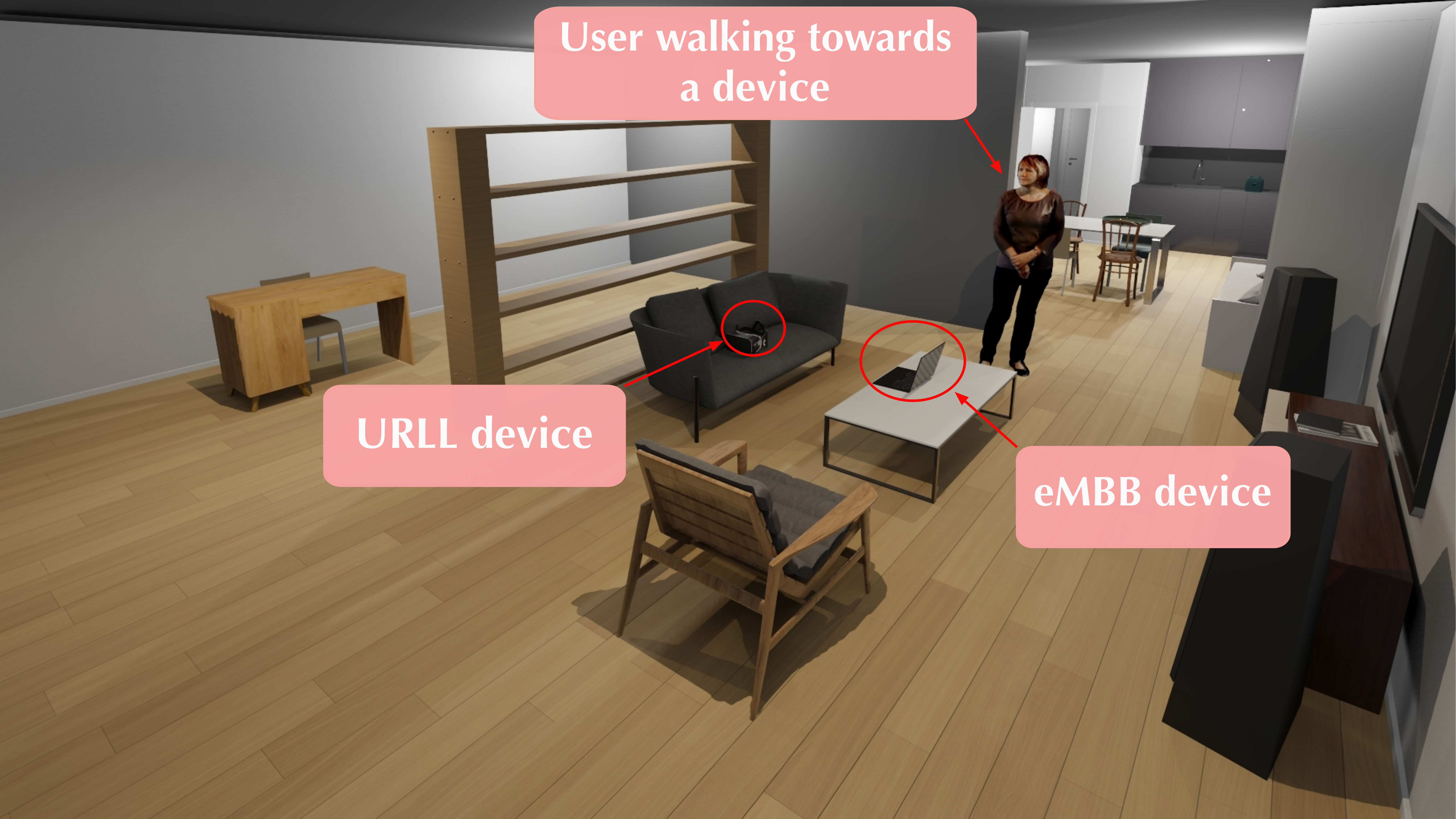}
		\caption{Camera 1 view}
		\label{fig:cam_1}
	\end{subfigure}
	\caption{The considered communication scenario. (a) shows the floor plan of the indoor environment, and (b) shows the perspective of camera 1.}
	\label{envi}
\end{figure}

\subsection{Communication Scenario and Development Datasets}\label{sec:scen_dataset}

\begin{figure*}
	\includegraphics[width=\linewidth]{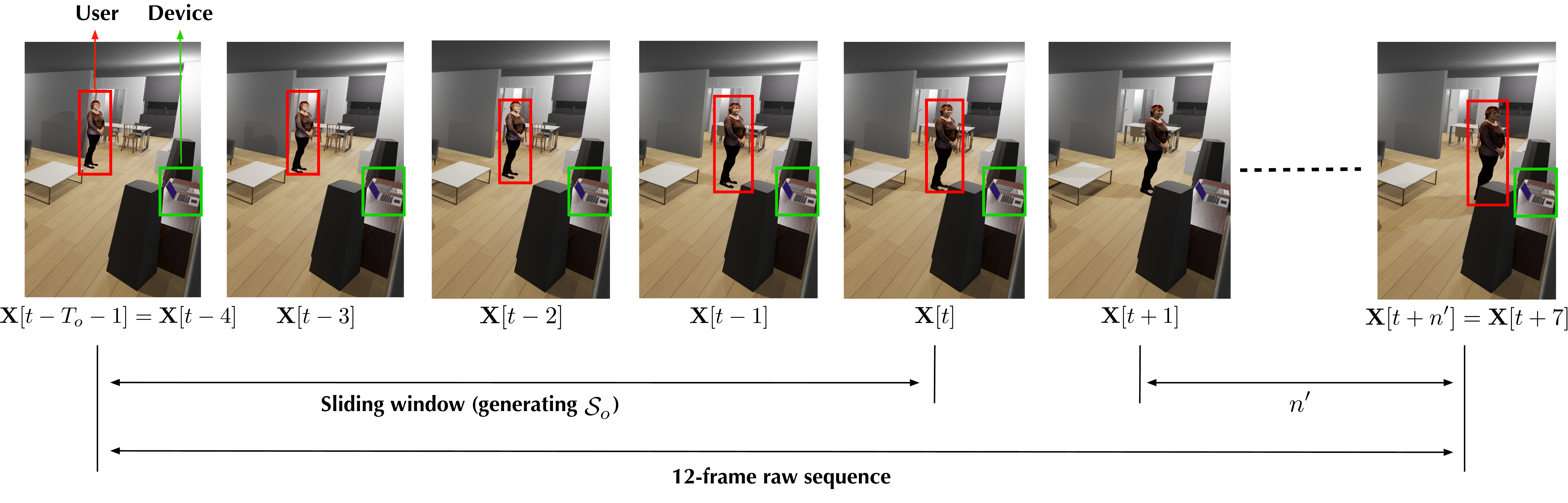}
	\caption{Example on how a 12-frame raw sequence is sliced using a $T_o=5$ window sliding across the sequence. The figure also illustrates how the request time is quantified as number of future frames $n^{\prime}$.}
	\label{fig:sample_example}
\end{figure*} 

The scenario considered in this paper embodies that described in Section \ref{sec:sys_mod}. It is posed in a home environment built using a game engine and following the framework of ViWi \cite{ViWi}. The scenario has two access points installed in two different rooms, see \figref{fig:scen}. Each one of the two caters to two different wireless services, a URLL and eMBB, within its designated room. Since service identification is all about proaction in wireless networks, the scenario is designed to have one human object moving towards a device that is dedicated to either URLL or eMBB. Different 6 human objects and two different devices are considered when building this scenario. For each choice of a human, a device, and a room, multiple trajectories for the human object going to the device are simulated to generate multiple video sequences. Each sequence depicts the human starting from a randomly selected point in the room and moving towards the device, which is also placed at different locations in each sequence. The result of the simulation is a group of raw variable-length video sequences that are henceforth called \textit{raw sequences}. They represent trajectories for various human, device, and room combinations.

Two development datasets are obtained from the raw sequences. One is a development dataset of short video sequences used to train and test the proposed DNNs on the service identification tasks. The other is a small object detector dataset used to train the object detector of the first DNN on detecting objects of interest. The details of the two dataset are given as follows:

\begin{itemize}
	\item \textbf{Service identification dataset:} short sequences of length $T_o = 5$ frames are generated by slicing the raw sequences using a sliding window that moves one frame at a time. Every 5-frame sequence (which is $\mathcal S_{o}$) is accompanied by two labels or targets, the type of service to be requested $y_{\text{type}}$ and its request time $y_{\text{rt}}$. All together, they make a 3-tuple or a single data point $(\mathcal S_o,y_{\text{type}},y_{\text{rt}})_u$ where $u\in\{1,\dots,U\}$ and $U$ is the total number of data points. The type of service is determined by the type of device that appears in the raw sequence, which is being sliced. In this paper, two services are considered, namely eMBB and URLL, and, therefore, they are awarded the following two labels, respectively, 0 and 1. On the other hand, request time is quantified in this work in terms of number of future frames $n^{\prime}$. Such number is defined as the number of remaining frames between the end of the observed sequence $\mathcal S_o$ and the end of the raw sequence from which $\mathcal S_o$ is being sliced. For instance, for a raw sequence of 12 frames, the first short sequence is obtained by slicing the first 5 frames, and the target request time $y_{\text{rt}}$ is 7 frames, see \figref{fig:sample_example}. Since the lengths of all raw sequences are finite, the number of future frames assumes values from a finite set of discrete numbers, i.e., $n^{\prime}\in\{1,\dots,T_p\}$ where $T_p$ is the largest possible number of frames between the end of a short sequence $\mathcal S_o$ and the end of its raw sequence. The final dataset contains approximately 36 thousand data points ($U = 36\times10^3$). This dataset is split $70-30\%$ to form the traning and validation sets, respectively.
	\item \textbf{Object detector dataset:} this is a small dataset of 200 frames selected at random from the raw sequences. The frames are manually annotated with bounding boxes of objects depicted in them and their class labels. The 200 frames are mainly used to get the object detector familiar with the objects of interests in the scenario, which is in-line with what is expected in real scenarios.
\end{itemize}

\subsection{Performance Evaluation Metrics} \label{sec:eval_met}
As the definition in Section \ref{sec:key_idea_def} states, service identification is a framework with two intrinsic tasks that are envisioned to guid a wireless network to proaction. As a result, any proposed solution under the service identification framework needs to be evaluated from two perspectives, the prediction quality of the solution and the wireless network performance using the solution. The next two subsections will highlight the main evaluation metrics that will be used to assess the quality of the proposed solutions.
\subsubsection{Machine learning} the first couple of metrics are concerned with the prediction quality of service type and request time, and, hence, they are considered machine learning metrics. Service type is posed as a binary classification problem for both proposed DNNs. Therefore, popular metrics that could be used here are precision, recall, and accuracy, \cite{Eval2011}. On the other hand, service request time is posed as a regression problem, which requires different kind of metrics. The choice in this paper is to go with the mean and standard deviation of the predictions. Due to the quantification of request time in the form of number of future frames $n^{\prime}$, the predictions are organized into groups based on the value of their groundtruth $y_{\text{rt}}$. In particular, for all data points with $n^{\prime}$ future frames, their predicted request times (i.e., $\hat y_{\text{rt,u}}$) are put together in one group. This results in $T_p$ groups, and for each one, the mean and standard deviation are calculated. Formally, this is expressed by
\begin{align}
	\bar{\mu}_{n^{\prime}} &= \frac{1}{U_{n^{\prime}}} \sum_{u = 1}^{U_{n^{\prime}}} \hat y_{\text{rt,u}}, \label{eq:mean} \\
	\bar{\sigma}_{n^{\prime}} &= \frac{1}{U_{n^{\prime}}} \sum_{u=1}^{U_{n^{\prime}}} ( \hat y_{\text{rt,u}} - \bar{\mu}_{n^{\prime}} )^2, \label{eq:std}
\end{align}
where $\bar{\mu}_{n^{\prime}}$ and $\bar{\sigma}_{n^{\prime}}$ are, respectively, the average and standard deviation of all predictions where the groundtruth targets are $n^{\prime}$, and $U_{n^{\prime}}$ is the total number of data points in the $n^{\prime}$-th group.

\subsubsection{Wireless network} The next set of performance evaluation metrics are tailored to the wireless network performance, more specifically reliability and utilization given respectively by \eqref{eq:rel} and \eqref{eq:eff}. The first step to show how those metrics are calculated in this paper starts with translating the predictions of a DNN into wireless terms. In particular, the service request time, which is measured in video frames $n^{\prime}$, should be converted into wireless slots to mark the start of the URLL packet, i.e., $x_k$. We do so by choosing $\text{FSR}= 33$ slots/frame where $\tau_v = 1/30$ sec/frame. This results in a wireless slot $\tau_w$ of approximately 1 millisecond. It is important to note here that given the dataset described in Section \ref{sec:scen_dataset}, there is only one URLL packet in every data point with $\hat y_{\text{type}}=1$. This means $k=1$ for all data points with a URLL device ($\hat y_{\text{type}}=1$) in the dataset and, therefore, the subscript in $s_k$, $x_k$ and $x^{\prime}_k$ will be henceforth replaced with $u$ to indicate the data point instead.

 Using the definition of FSR, we move on to show how reliability and utilization are calculated. We consider a guard band of $T_G$ slots from both sides of the predicted request time of a URLL service, where $2 T_G+1$ is defined as the amount of slots kept free in anticipation of URLL packet transmission. 
For the $u$-th data point, if $\hat y_{\text{type}}=1$, we check the predicted request time $\hat y_{\text{rt}}$ and convert it into wireless slots using the FSR and $n^{\prime}$. The result of the conversion is the predicted beginning of the URLL packet $\hat{x}_u$. 
Based on the predictions $\hat y_{\text{type}}=1$ and $\hat{x}_u$, the slots $[\hat{x}_u-T_G, \hat{x}_u+T_G]$ are left empty, i.e., not utilized by the eMBB service to prevent any possible collision. We assume that only a single prediction is made for each packet by the neural network. We denote the event that the beginning of the predicted URLL packet of the $u$-th data point is within the guard interval by $\mathcal E_{t,u}= \{x_u \in [\hat{x}_u-T_G, \hat{x}_u+T_G]\}$ and we define its compliment as $\mathcal E_{t,u}^c$. We also denote the event of correct URLL service prediction of the $u$-th data point by $\mathcal E_{s,u} = \{\hat y_{\text{type}} = y_{\text{type}} \cap y_{\text{type}}=1\}$. Then, we can formulate the reliability per data point as
\begin{equation}\label{eq:rel}
	R_u =  \mathbbm{1}{\{ \mathcal E_{s,u}\cap \mathcal E_{t,u} \}},
\end{equation}
where $\mathbbm{1}$ denotes the indicator function of the case given between the curly braces. Similarly, we can write the utilization defined in \eqref{eq:eff} for a single data point, after the completion of the transmission. Recall that the number of the slots that are not kept idle defines utilization. The number of idle slots is determined by two events: 1) If a packet arrival is predicted, but the arrival is not within the guard interval, $2 T_G+1$ slots are not utilized. 2) If the arrival is within the guard interval, the slots after the start of the guard interval and before the start of the transmission are not utilized. Assuming the guard intervals of two predictions do not intersect, we can write the utilization per data point as

\begin{align}
	Z_u &= 1 - \frac{\sum_{q=1}^{t_w} (1-z_q)}{t_w} \label{eq:utilinv} \geq 1 - \frac{\sum_{q=1}^{t_w} (1-z_q)}{\rho}
	\\ &= 1 - \frac{1}{\rho} \bigg(\mathbbm{1}{\{ \mathcal E_{t,u}^c \cap \mathcal E_{s,u} \}} (2 T_G+1) \notag \\ & \quad\quad + \mathbbm{1}{\{ \mathcal E_{t,u} \cap \mathcal E_{s,u} \}} (x_u - \hat{x}_u + T_G)\bigg) \label{eq:util}
\end{align}
where 
 \eqref{eq:util} is obtained using the event definitions for a single data point, and $\rho$ is a constant such that $t_w \geq \rho$. For the derivation to be accurate, the intersection of the guard intervals should not occur. Hence, the selection of $\rho$ needs to cover the sample duration along with the addition of a guard interval, i.e., $\rho \geq T_G + T_p$. In addition, $t_w$ can be interpreted as the frequency of the packet transmissions since a single wireless packet is transmitted within that duration, with a maximum arrival frequency. 

We next write the expected reliability and utilization by assuming independence of all the variables and packets:
\begin{itemize}
	\item \textbf{Reliability calculation:} Formally, the expected reliability can be given by
\begin{equation}\label{eq:exp_rel}
	ER \% = \mathbb P\left( \mathcal E_{t,u}, \mathcal E_{s,u} \right) \times 100.
\end{equation}
\eqref{eq:exp_rel} is translated in our evaluation into averaging over the validation data points. Therefore, \eqref{eq:exp_rel} is expressed as follows
\begin{align}\label{eq:rel_metric}
	ER\% &=  \left(\sum_{u=1}^{U_{\text{urll}}}\frac{\mathbbm{1}\{\mathcal E_{s,u} \cap \mathcal E_{t,u}\}}{U_{\text{urll}}}\right)\times 100,
\end{align}
where $U_{\text{urll}}$ is the total number of data points in the validation set where a URLL service is requested.

\item \textbf{Utilization calculation:} The expected utilization, on the other hand, can be written as
\begin{align}\label{eq:exp_util}
\begin{split}
&EZ\% = 1 -  \frac{1}{\rho} \bigg(\mathbb{P}{\{ \mathcal E_{t,u}^c, \mathcal E_{s,u} \}} (2 T_G+1) + \mathbb{P}{\{ \mathcal E_{t,u}, \mathcal E_{s,u} \}} \\ & \quad\quad \sum_{h \in \mathbbm{Z}^+} \mathbb{P}(x_u-\hat{x}_u=h| \mathcal{E}_{t,u} \cap \mathcal{E}_{s,u})(h + T_G)\bigg) \times 100.
\end{split}
\end{align}
Although the parameters in \eqref{eq:exp_util} can be derived for a given activity and packet length distributions, our simulations are based on the validation set, and, hence, \eqref{eq:exp_util} is computed by averaging $Z_u$ over all URLL samples in our dataset. This could be written as
\begin{equation}
	EZ\% = \bigg( \frac{1}{U_{\text{urll}}} \sum_{u=1}^{U_{\text{urll}}} Z_u \bigg)\times 100.
\end{equation}

\end{itemize}

\subsection{Training of DNNs}
The proposed solutions adopt two different design perspectives, and hence, the two DNNs have different training procedures and hyper-parameters. The following two subsections detail that.
\subsubsection{Two-stage}
the two-stage architecture is trained in two back-to-back phases. The first one trains the object detector to recognize the objects of interest while the second trains the GRU network of the sequence-based decision stage. For the first phase, instead of designing and training a detector from scratch, a more efficient approach based on transfer learning is followed. It utilizes a popular detector trained on a large dataset. The detector of choice in this work is a YOLO \cite{Yolo} model trained on the COCO dataset \cite{coco}. Such model has been shown to be fast and to achieve reliable detection performance compared to other models \cite{Yolov2}. The training hyper-parameters presented in \cite{Yolov2} are used to finetune the YOLO model on the object detection dataset in Section \ref{sec:scen_dataset}. The second training phase starts once the YOLO model is trained. It uses the service identification dataset and relies on the predictions of the YOLO model. The GRUs in this phase learn to predict the service request time since the service type is based on majority voting, as described in Section \ref{sec:2_stage}. Table \ref{table:hyper} lists the main hyper-parameters. 

\begin{table}
	\centering
	\caption{Training Hyper-parameters}
	\begin{tabular}{c|cc}
	\hline\hline
	\multirow{8}{*}{GRU training}
	& Hidden state dimension & 64 \\
	& Number of GRU layers & 2 \\
	& Dropout percentage & 25$\%$ \\
	& Solver & Adam \cite{Adam} \\
	& Learning rate & $1\times10^{-3}$ \\
	& Maximum number of epochs & 50 \\
	& Learning rate schedule & 0.1 @ epoch 20 \\
	& Mini-batch size & 100 \\
	\hline
	\multirow{7}{*}{Convolution training}
	& Solver & SGDM \cite{DLBook} \\
	& Learning rate & $1\times 10^{-4}$ \\
	& Weight decay & $1\times 10^{-5}$\\
	& Momentum & 0.9 \\
	& Maximum number of epochs & 6 \\
	& Learning rate schedule & 0.1 @ epoch 5 \\
	& Mini-batch size & 100 \\
	\hline\hline
	\end{tabular}
	\label{table:hyper}
\end{table}

\subsubsection{End-to-end}
The end-to-end architecture is trained on the service identification dataset presented in Section \ref{sec:scen_dataset}. As its name suggests, the architecture is trained end-to-end once for service type prediction and the other for service request time. The only difference between the two runs is the prediction layer. This layer is a two way classifier in the case of service type prediction whereas it is a single-neuron regressor in the case of request time prediction. The hyper-parameters of the training process are listed in Table \ref{table:hyper}. These parameters are the same of for both tasks. The implementation details of this are available at \cite{myGithub}.

\section{Performance Evaluation}\label{sec:perf_eval}
The performances of the two proposed architectures are evaluated using the development datasets and the evaluation metrics introduced, respectively, in Sections \ref{sec:scen_dataset} and \ref{sec:eval_met}. The subsections below lay out a detailed analysis of their performances. They start with analyzing the machine learning performance and, then, put that analysis into wireless network terms.

\subsection{Service Identification Performance} \label{sec:serv_id_perf}
\begin{figure}[t]
    \centering
    \begin{subfigure}{0.45\textwidth}
        \centering
        \includegraphics[width=0.57\columnwidth]{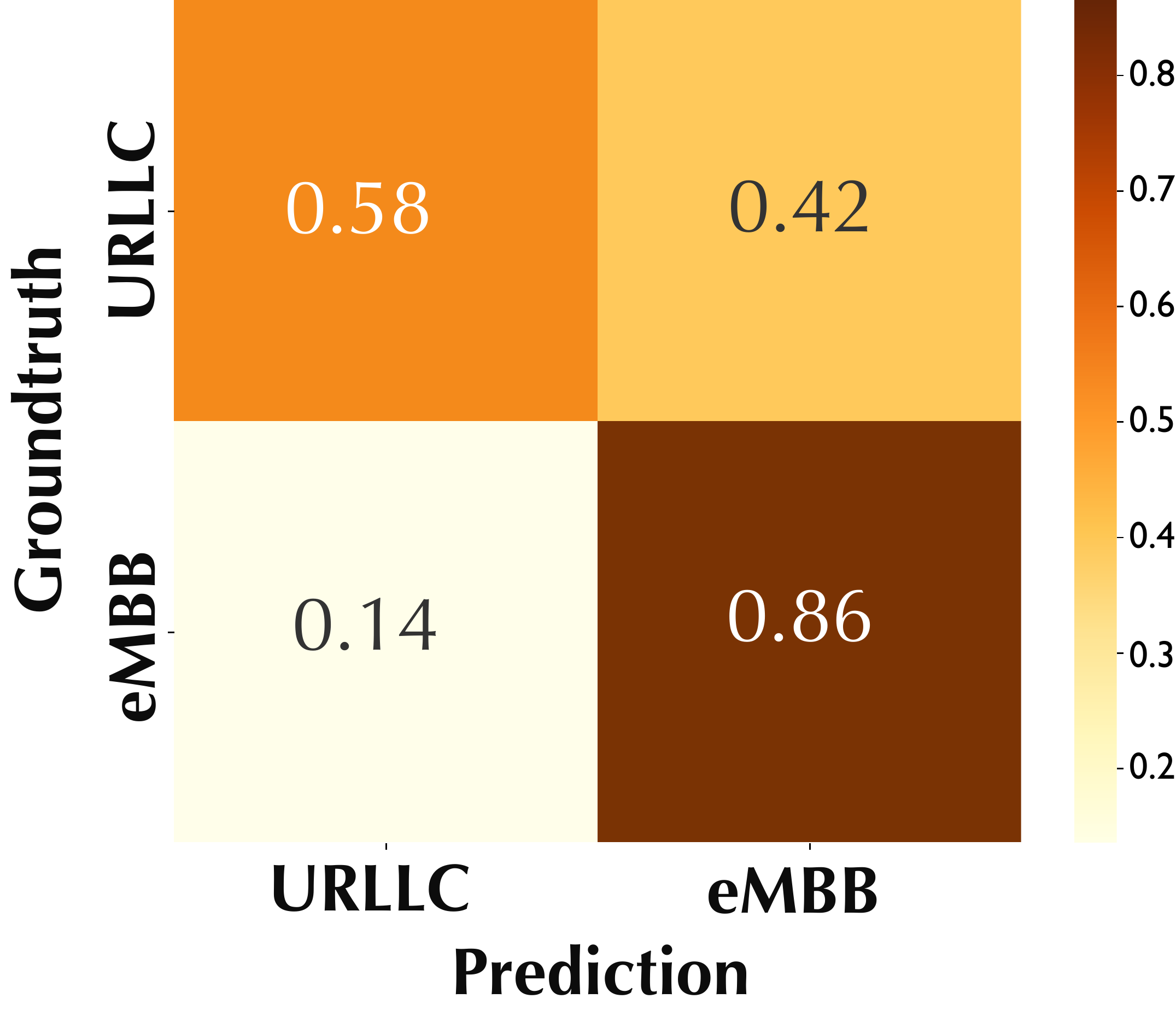}
		\caption{}
		\label{fig:2_stage_cm}
    \end{subfigure}
    \begin{subfigure}{0.45\textwidth}
        \centering
        \includegraphics[width=0.57\columnwidth]{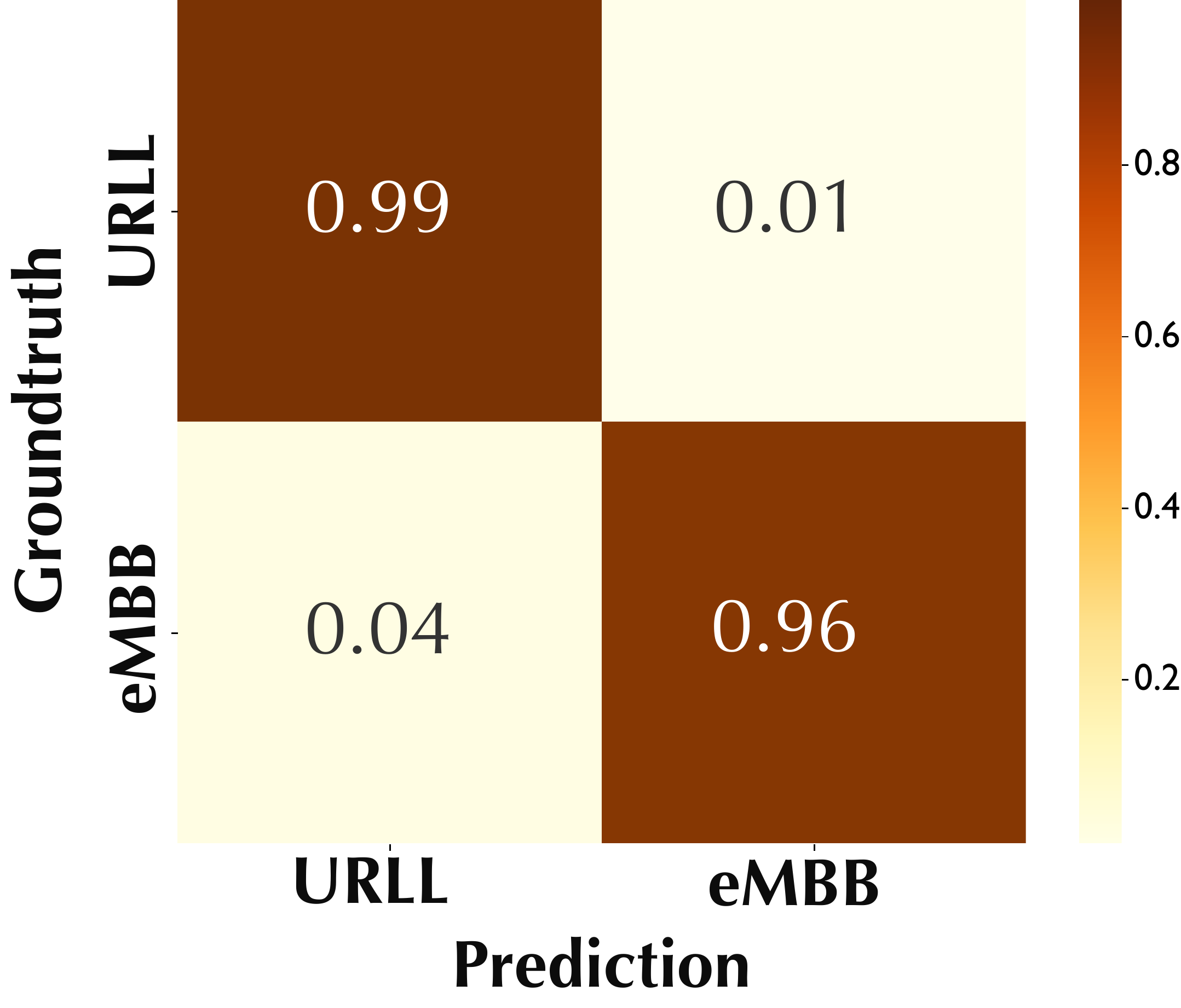}
		\caption{}
		\label{fig:ete_cm}
    \end{subfigure}
    \caption{The confusion matrices of the two proposed DNN solutions. (a) shows the matrix for the two-stage architecture while (b) shows the matrix for the end-to-end architecture.}
    \label{fig:conf_mat}
\end{figure}

The core tasks of service identification are predicting service type and predicting its request time. In the validation set, there is a total of 5492 data points with URLL service and 5558 data points with eMBB service. The analysis of the machine learning performance starts with the task of predicting service type. \figref{fig:conf_mat} shows the confusion matrices of both architectures. What could stand out from the first glance are the askew performance of the two-stage architecture in favor of eMBB and the healthy balanced performance of its counterpart across both service categories, i.e., the true negative rate is much higher than the true positive rate for the two-stage architecture. This skewness in the two-stage performance could be traced back to the objects detector. It clearly indicates that the detector is struggling in identifying the URLL device, which results in the reduced true positive rate. This trend is confirmed by the precision performance, which reflects how trustworthy the URLLC predictions are \footnote{Precision is defined as the ratio of number of true positive detections to the sum of true positive and false positive detections.}. Two-stage achieves a little over $80\%$ precision for a recall rate (true positive rate) of $58\%$. This does not bode well for the architecture especially when compared to the impressive $98\%$ precision at $99\%$ recall of the end-to-end architecture. However, it points out to an interesting result; the two-stage struggles in predicting URLLC, yet when it does, its predictions are $80\%$ accurate.

\begin{figure}[t]
\centering
\begin{subfigure}{0.45\textwidth}
\centering
\includegraphics[width=1\columnwidth]{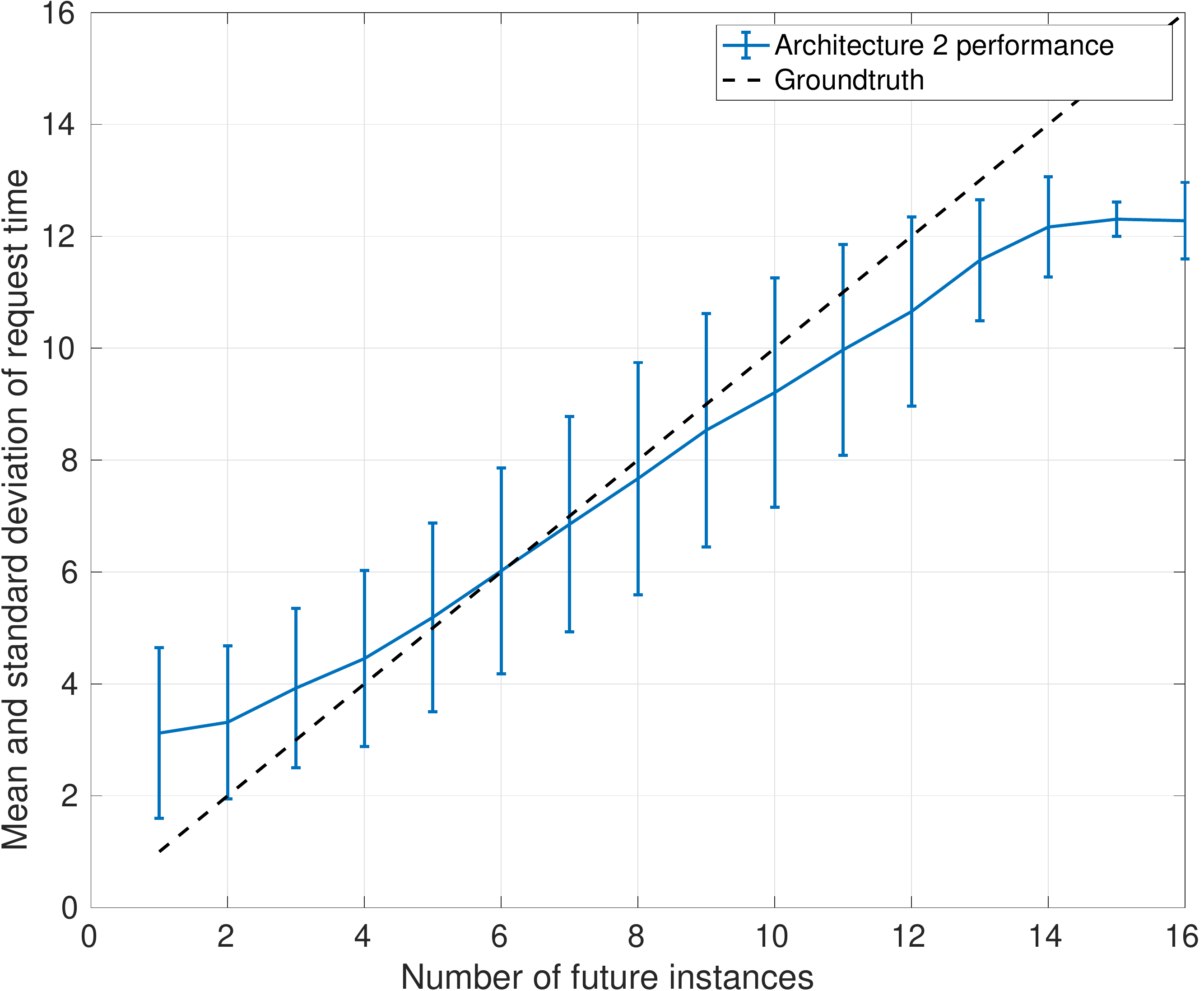}
\caption{}
\label{fig:2_stage_perf}
\end{subfigure}
\begin{subfigure}{0.45\textwidth}
\centering
\includegraphics[width=1\columnwidth]{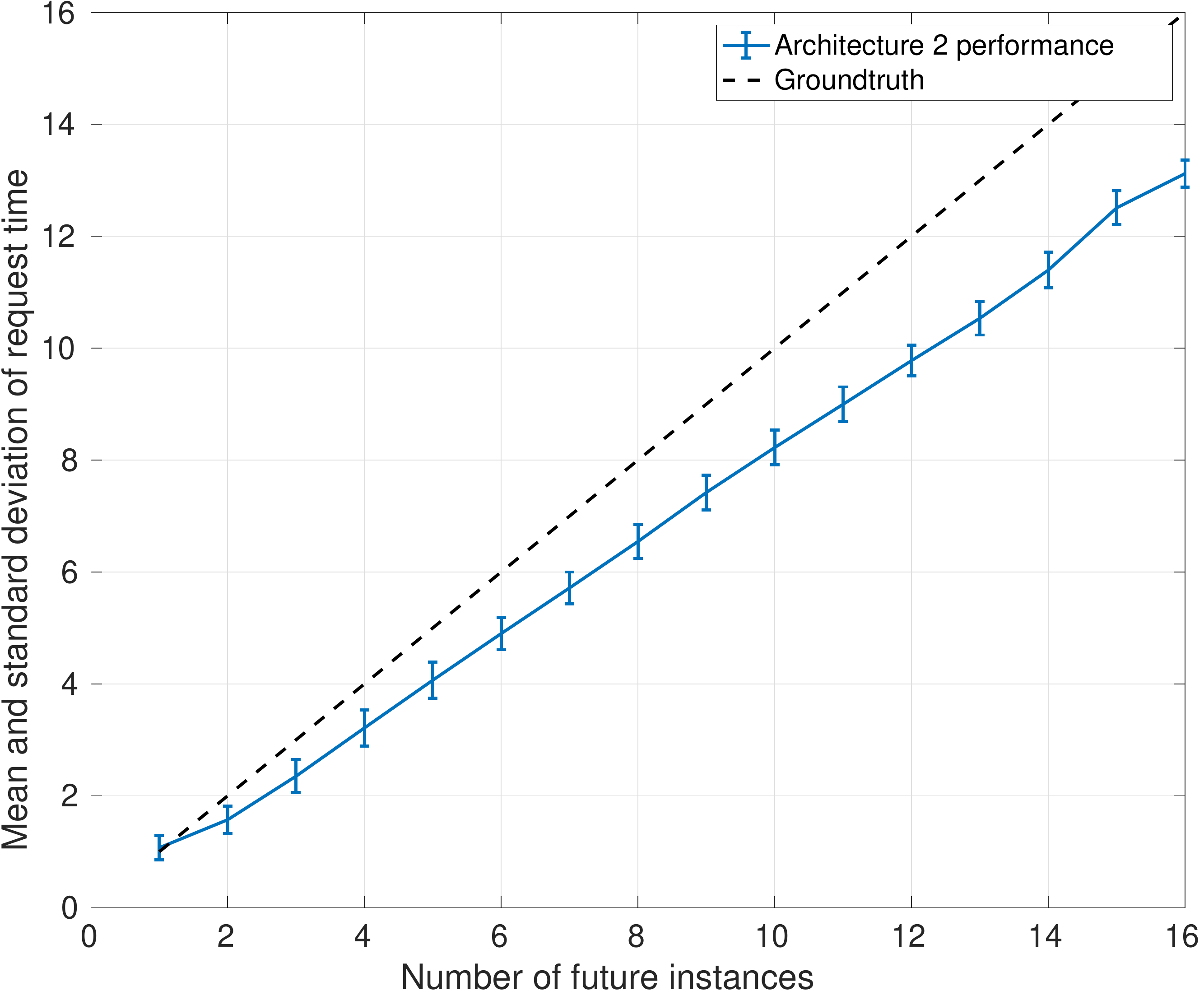}
\caption{}
\label{fig:ete_perf}
\end{subfigure}
\caption{Mean and standard deviation of service request time for both proposed DNNs. (a) shows the performance of the two-stage architecture while (b) shows the performance of the end-to-end architecture.}
\label{fig:rt_perf}
\end{figure}

\begin{figure}[t]
\centering
\begin{subfigure}{0.45\textwidth}
\centering
\includegraphics[width=1\columnwidth]{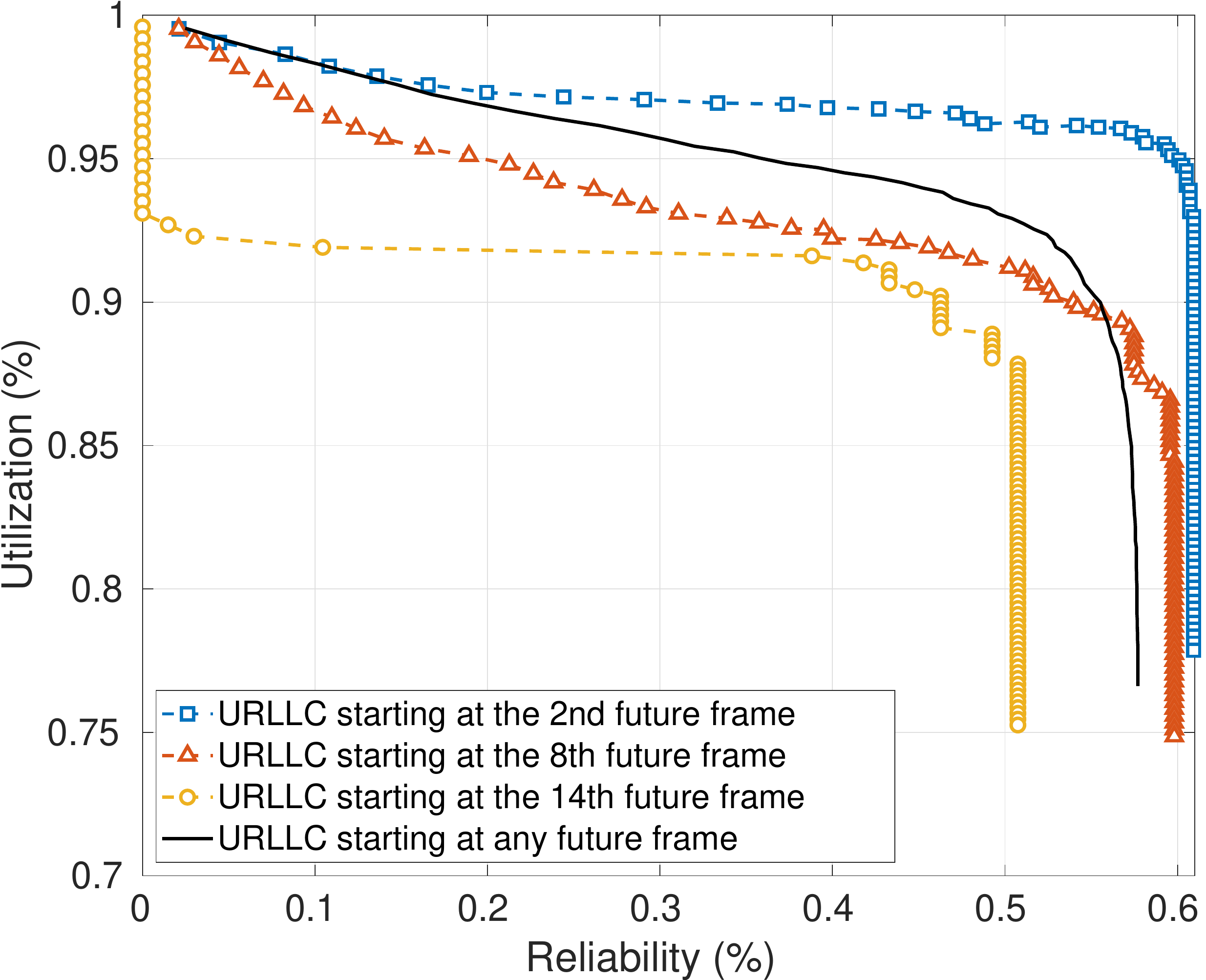}
\caption{}
\label{fig:2_stage_rVSu}
\end{subfigure}
\begin{subfigure}{0.45\textwidth}
\centering
\includegraphics[width=1\columnwidth]{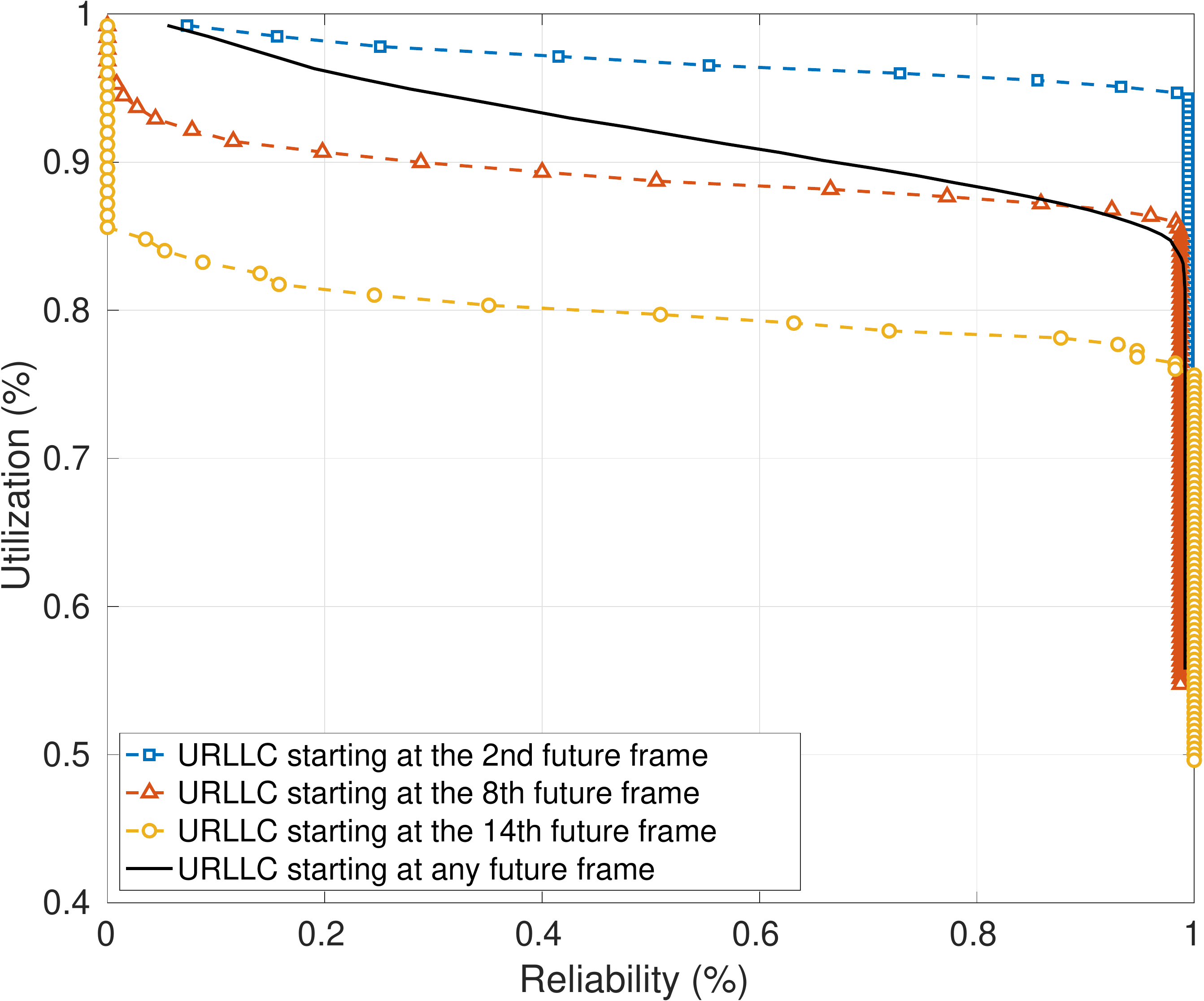}
\caption{}
\label{fig:ete_rVSu}
\end{subfigure}
\caption{The spectral efficiency versus reliability performance of a proactive wireless network employing one of the two proposed DNNs. (a) is for the two-stage architecture while (b) is for the end-to-end architecture.}
\label{fig:eVSu}
\end{figure}

For the second task in service identification, which is service request time, the two-stage is lagging behind the end-to-end architecture, as well. \figref{fig:rt_perf} shows the mean and standard deviation of the predicted service request time for both architectures, calculated using \eqref{eq:mean} and \eqref{eq:std}. The two-stage architecture again shows a fluctuating average performance across the predication range; for services that start in the near future (between 1 and 4 future frames) or those that start in the far future (between 12 and 16), it tends, respectively, to over- and under-estimates the request time. Only when the request time is within the range of 4 to 11 future frames does the architecture achieve a relatively linear average prediction performance. Such region could be labeled ``trustworthy'' if it was not for the the increased standard deviation. This unstable performance is conjectured to be the consequence of how the sequence-based decision stage operates; it relies on bounding box information to learn service time, which is very limited in terms of what it encodes. For instance, bounding boxes could be good indicator of speed and direction when the motion of the human subject is majorly happening along the x-y plan of the image, but when both speed and direction have major components along the z-axis (depth), bounding boxes are expected to be less informative. 

In contrast, the end-to-end architecture performs significantly better in service request time. It consistently under-estimate the request time but with an almost linearly-increasing error rate with respect to the number of future frames, and it exhibits an almost constant standard deviation for each value of future frames. The former has the advantage of being easily corrected with a scaling factor while the latter could be weathered by the careful design of guard bands ($T_G$). This good performance, in general, could be attributed to the fact that the architecture is trained to learn tailored intermediate features to the task in hand as opposed to being forced to produce some hand-crafted features such as bounding boxes. More information on the power of learning features versus engineering them could be found in \cite{DLBook,LearnDeepArch,DL:meth_App}.

\subsection{Network Reliability and Spectrum Efficiency}
With the machine learning performance in mind, the network reliability and efficiency when both architectures are deployed is studied in this subsection. \figref{fig:eVSu} shows two efficiency versus reliability plots obtained by varying the choice for a guard interval, namely $T_G\in\{0.1, 0.2, 0.3,\dots,10\}$ frames. The struggles of the two-stage architecture in evenly predicting the service type is reflected in the average performance in \figref{fig:2_stage_rVSu}. 
It is clear how the efficiency rapidly degrades when the reliability is increased beyond $50\%$. This could be traced back to the low true positive rate (recall) of the architecture, which impacts the numerator of \eqref{eq:rel_metric}. On the other hand, the high positive rate of the end-to-end architecture produces a more consistent and trustworthy network performance. The average reliability could be pushed as close to $100\%$ as possible with a degradation of around $15\%$ in average efficiency. In an operational sense, this means that a wireless network can effectively allocate $85\%$ of the URLL slots to an eMBB service with a $1\%$ likelihood of collision.

The above efficiency and reliability performances are, overall, quite intriguing. This could be rooted in two important reasons: (i) they paint a picture of an auspicious framework for proaction and (ii) they are achieved with almost zero communication burden on the wireless network. Whether it is the two-stage architecture or its more successful counterpart, it is critical to note that the performances above are obtained with proaction. They both encourage further investigation into how deep learning (or machine learning in general) could be used with computer vision to augment resource allocation in wireless networks. The two, deep learning and computer vision, have the advantage of not adding extra communication burden on the network. In fact, they make the most out of what is already there, and this is what the second reason is emphasizing. In general, a service identification solution could be viewed as a transparent addition to the wireless network, i.e., the kind of addition that may not require an overhaul of the existing infrastructure.

\section{Conclusion}\label{sec:concl}
This work presents an argument for computer vision and deep learning as enablers to proactive resource allocation in wireless communication networks. The argument is posed in the form of a framework termed service identification, in which a machine learning algorithm learns to identify the type of an incoming service and its request time using visual data (e.g., RGB video frames). The framework is studied in an indoor wireless network scenario with two coexisting services, a URLL and an eMBB. Two algorithms based on deep neural networks, called two-stage and end-to-end, are developed to predict incoming services and their request time. The two algorithms are shown to help the wireless network achieve different levels of proaction, which are measured using service reliability and time-frequency resource utilization. The two-stage algorithm achieves a maximum of $\sim 78\%$ utilization at a reliability of $\sim 60\%$ while the end-to-end achieves $85\%$ utilization at a $\sim 98\%$ reliability. The big difference in the reliability performance could be attributed to the sequential architecture of the two-stage algorithm; the learning failures of the first stage limits what the second stage can learn. Such problem does not exist in the end-to-end algorithm. Despite their varying performances, the two algorithms are good demonstrations for the potential of service identification. They both achieve those utilization and reliability performances by only utilizing visual data. This is an important point to emphasize here because in reality, service identification is not expected to operate in isolation from the rest of the wireless network components. On the contrary, it is envisioned as an extra layer of intelligent processing that could work hand-in-hand with classical means of resource allocation to meet the seemingly conflicting reliability, latency, and utilization demands of heterogenous services in future wireless networks.

\balance

\end{document}